# A Novel Nudity Detection Algorithm for Web and Mobile Application Development

Rahat Yeasin Emon
*Department of Computer Science and Engineering*
*Chittagong University of Engineering and Technology*
*Chittagong-4349, Bangladesh*
emon13111@gmail.com

*Abstract*— In our current web and mobile application development runtime nude image content detection is very important. This paper presents a runtime nudity detection method for web and mobile application development. We use two parameters to detect the nude content of an image. One is the number of skin pixels another is face region. A skin color model based on RGB, HSV color spaces are used to detect skin pixels in an image. Google vision api is used to detect the face region. By the percentage of skin regions and face regions an image is identified nude or not. The success of this algorithm exists in detecting skin regions and face regions. The skin detection algorithm can detect skin 95% accurately with a low false-positive rate and the google vision api for web and mobile applications can detect face 99% accurately with less than 1 second time. From the experimental analysis, we have seen that the proposed algorithm can detect 95% percent accurately the nudity of an image.

*Keywords—pornography, nudity, non-nudity, skin-color-model, skin-detection, face-detection*

## I. Introduction

Pornographic images in the web and mobile applications have become a great concern in our society. Nude images are offensive for some kind of people [1]. It can be used as some type of crime, such as child pornography [2]. Pornographic images affect children's and teenagers' mental and physical health. Parents are afraid and discourage their children to use the internet. Again pornographic content often contains misleading information such as malicious threats.

In today's internet world videos and images are the most important phenomenon. The vast development of web and mobile applications, it is often affected by pornographic content. To protect the internet from pornographic image content, theirs need a nudity checker.

Filtering an image to detect whether it contains nude content is a vast research area. Often image nudity is very difficult to be described [6]. Different people have different perceptions about which images nude which are not. Machine learning is the most reliable technique for image processing. But machine learning technique needs a huge amount of data set and time. Concerning space and time currently available method to detect nude image is still crude. Technology companies demands new pornography detection method.

In our modern era, web and mobile applications are mostly used technology. Internet images often uploaded from the web and mobile devices. In our internet server-client model, to protect server busy theirs need a fastest front-end validation technique which can protect the nude images being uploaded into an internet server.

The paper presents a runtime nudity detection technique for web and mobile application development. To detect nudity two important parameters of the human body are used one is skin color another is face region. Raw RGB and HSV color spaces are used to detect skin color. Google vision api or google firebase ML kit is used to detect face region. By the ratio of total number of skin pixels and the number of skin pixels in the face region, we detect whether the input image contains nude content or not.

## II. Background

To detect the nudity of an image there exists several theoretical and practical research work they are mainly based on body-shape or object-detection, detection of skin color and pre-classified database is used [1]. In [3,9] only skin pixels are considered to detect nudity. A skin color model based on RGB, Normalized RGB, HSV, and YCbCr is used to detect skin color pixels. By the percentage of skin color pixels and skin regions, an image is classified nude or not. The disadvantage of this method is that it does not classify images containing a large amount of skin color that does not have any nudity for example images containing only faces have a high ratio of skin color for the whole image and this method mistakenly identified those images as nude image.

An updated version of paper [3] is discussed in [6,7]. They proposed machine learning-based training and learning rule for nude image classification. However, machine learning-based nudity detection is very time consuming and will not discuss in this paper.

In [4,5] skin color region is detected and then the detected regions are converted into body shape structures to detect nudity of image. In [8], the body shape was extracted and fed into a classifier based on skin color detection it decides whether or not the body-skin regions represented nudity. Fu and Wang [10] proposed a method where ROI (Region of Interest) is located by identifying skin-like pixels using YCbCr color spaces. Then ROI is classified into SVM classifiers to detect the pornographic image. The Yang [11] method also determines the ROI with skin information. The image was classified into different regions and points. Based on the ROI this method obtains some features for image classification. In [12], the property of an image such as color and texture is obtained and then a pattern classifier is used to detect nudity. However, the body-shape method discussed in this paragraph failed to detect nudity in images with facial content, partial explicit content, and a high level of illumination. To resolve these issues in this paper we separately calculate the face regions.

In [13], a pre-classified database is used to match an image. When the number of pre-classified images from the database reaches a threshold for an image, then it is labeled as a nude image. This pre-classified database image detection cannot be used in real-time applications such as web and mobile applications front-end validation.

In [14], face-body based nudity detection is used where face and body are mandatorily detected first with object shape detection method then compare it with a threshold that is relative to the face-body ratio of the current body area. Kosesoy [15] proposed a face based nudity technique where a ratio is obtained from dividing the total number of nude pixels by the total number of faces. In this study, after detecting faces body region calculated for each face region according to the ideal face-body ratio [16]. However, this method cannot detect the nudity in non-face images and the face-body ratio is not the ideal ratio to detect nudity.

The face-body skin ratio based nudity detection is discussed in [17]. In this study, they detect face bounding box and body skin pixels. The ratio of the face bounding box and body skin pixels determine the image nudity. In the case of skin detection and nudity detection, this method provides low accuracy results.

In our nudity detection method, we are focusing on two-parameter of human body one is detecting skin region and another is face region. In this algorithm, the face region is not mandatory. But if face exists it gives a higher accuracy to detect nudity of an image. By the percentage of skin region pixels, face regions skin-pixel, and the total pixels of an image we detect whether the image contains nudity or not.

A. Skin Detection

Skin color detection is our key point to detect the nudity of an image. To detect the skin region, we have used a combination of research methods. We proposed a new skin detection model which is based on Kolkur [18] and Osman [19] skin color model. Now will see few terminologies related skin detection.

*A.1. Skin color*

Skin is the largest organ of the human body [20]. Skin color is our key point to detect the nudity of an image. Color is lightweight and inexpensive therefore it is suitable for real-time object characterization [21]. The nude image has a large number of skin regions, so skin color is the basic feature to detect nudity. Today skin detection is used in face detection, nude image detection, human detection, and hand gesture analysis [19]. The color-based skin detection has a disadvantage that it will not work in white and black images. However, back an white images are rarely existed in our web and mobile application system.

Skin color is produced from melanin, hemoglobin, carotene, and bilirubin. Hemoglobin gives the blood a reddish color or bluish color while carotene and bilirubin give skin a yellowish appearance [19]. The lightness or darkness of skin depends on the amount of melanin in the skin [3].

The simplest methods in skin detection assume skin color have a certain range in a color space [3]. Different color spaces are used to detect skin. The study [22] indicates that skin tones have similar in chrominance value and they differ mainly in their intensity value. The researcher Bosson [23] and Chan [24] agreed that the color spaces selection is not so critical if proper training and data set are used.

To detect skin pixel, we need a decision rule to identify whether a color image pixel is skin or not. Decision rules to detect skin pixels can be simple threshold-based color spaces [25,26] or can be complex as neural networks [27], Bayesian [28], maximum entropy [29] or k-means clustering [30]. The skin color boundary method in certain color spaces is the most popular technique to detect skin because it is lightweight and easy to implement [3-7].

*A.2. Color Spaces*

Color spaces are a tool for representing color information usually consists of three or four color information. It is a system where each color represented by a single point. Different types of Color spaces are used in computer graphics, image processing, TV broadcasting, and computer vision [18]. One-color space is appropriate in one application might be inappropriate in other applications. Color space selection is the primary process to detect skin color model and classification. RGB based color spaces (RGB, Normalized RGB), Hue based color spaces (HIS, HSV an HSL) and luminance based (YCbCr, YIQ and YUV) are popular color spaces for skin color modeling [18]. In our proposed algorithm we used RGB, and HSV color space to detect skin pixels.

*A.3. The RGB Color Spaces*

The RGB color space is the default color space in digital image processing. Any other color spaces can be generated from this RGB spaces [3]. This color spaces have three basic colors Red, Green, and Blue. All color is the mixture of these three base color. Every color has Red, Green, and Blue color component. In RGB color spaces Red, Green, and Blue values are ranging from 0 to 255 (256 different value). It makes 256 * 256 * 256 = 16777216 possible unique colors. However, this RGB color space cannot alone classify skin for that case Normalized RGB, HSV are also used to generate a skin detection system.

*A.4. The Normalized RGB*

The Normalized RGB is another method to detect skin pixels. It removes the image luminance through normalization. The normalized Red(r), Blue(b) and Green(g) obtained after dividing the component by summation of RGB value.

$$r = R / (R+G+B)$$
$$g = G / (R+G+B)$$
$$b = B / (R+G+B)$$

*A.5. The HSV color spaces*

The RGB color space is not so reliable for human skin interpretation. Researchers use hue, saturation, and brightness to describe human body color objects [3]. Here, hue describes an angle between 0 to 300 degrees, saturation describes the grayness ranging from 0 to 100%, and the value represents the brightness or intensity of color ranging from 0 to 100%.

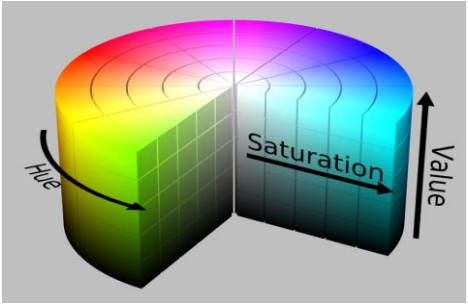

*B. Face Detection*

Face detection is our second parameter to detect image nudity. In this paper, we will discuss the practical nudity checker implementation for web and mobile applications. We used google-vision face api to detect face region.

*B.1. Google Vision Face Api*

Google vision face api is a powerful tool for face detection for web and mobile application development where powerful machine learning tool is used [32,33]. This api is lightweight and easy to implement. For the implementation purpose of web and mobile applications here, we used this face api to detect the bounding box of the face region.

*B.2. Cloud Vision Face Api for Web Application*

Google cloud vision face api for web application is lightweight, reliable, and it is easy to implement. For the web application, it is not totally free.

*B.3. Face Api for Mobile Application*

Android and IOS are the two popular mobile application platform. Google has a mobile vision face or ML kit api to detect faces region in android and ios based device application system. For device-based api, the face detection system is free.

### III. THE ALGORITHM

In our nudity detection system, we detect total skin pixels, face region, and skin pixels in the face region of an image. There are many skin color model rule exists, there are mainly based on Kovac rule [31], Swift rule [39], Saleh model [40], and Kolkur method [18]. We experimentally analyze every method, after combining those methods here we propose a skin detection rule for web and mobile applications. Our Skin detection rule:

Detect skin using RGB, HSV color spaces, Here (R) for Red, (G) for Green, (B) for Blue, (H) for Hue, and (S) for Saturation value.

If $R > 95$ and $G > 40$ and $B > 20$ and $R > G$ and $R > B$

and $|R - G| > 15$ and $0.0 <= H <= 50.0$

and $0.23 <= S <= 0.78$

then that pixel is **skin pixel**.

After the detect skin, we detect the face region. We used google vision tool to detect face regions. Now the overall algorithm works in the following manner.

1. Scan the image and get the pixels of the image.
2. Detect face regions if it exists.
3. Check every pixel whether it is skin pixel or not.
4. Calculate the number of skin pixels in that image.
5. Calculate the skin pixels in the face regions.
6. If the face region exists
    a. If (number of face skin pixels / total number of skin pixels) < 0.15 flag the image as **nude image**.
    b. Else flag the image as **non-nude image.**
7. Else no face region exists
    a. If (number of skin pixels / total number pixels of that) > 0.38 flag the image as **nude image**.
    b. Else flag the image as **non-nude image.**
8. End.

Now, we will discuss the practical demonstration of the algorithm.

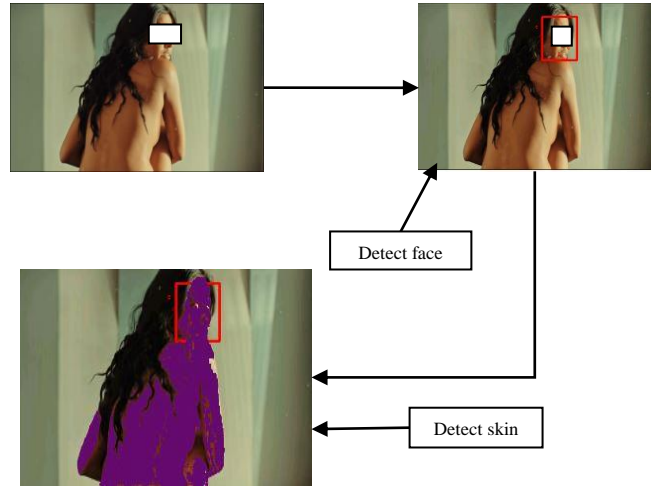

Fig. 1. The proposed algorithm, detect face and skin pixels

The above figure describes the procedure of our algorithm. The skin pixels depicted as pink color. In our nudity detection algorithm, we first detect face region (if exists) then detect skin pixels of face region and the total image area. If face region exists, then the ratio of face skin region and the total skin region detect the nudity of the image. If the ratio is above a threshold value, then we flag that image as nude image. Again if the face region does not exist, then ratio of total skin pixels and total image pixels determine the nudity of that image.

### IV. RESULTS

We used separate data sets to test our skin detection algorithm and nudity detection algorithm. For skin detection, we take 5,582,768 manually labeled skin pixels and 30,724,928 manually labeled non-skin pixels. From this testing we have seen that our skin pixels' detection algorithm detect skin 94.89% with 7.24% false positive rate. The following table we will show a comparison between our proposed skin detection rule with some popular skin detection model.

TABLE I. COMPARISON OF SKIN COLOR MODELS

| Color Model | Correct Detection | False Positives |
|---|---|---|
| Kovac | 10 | 42.5 |
| Solina | 98 | 14 |
| Ghazali | 94.91 | 33.07 |
| Rigan | 94.32 | 5.98 |
| Proposed skin color model | 94.89 | 7.24 |

Again, for testing nudity detection we used 435 nude images and 345 non-nude images. From that testing, we have seen that the proposed nudity detection algorithm can 97% accurately detect the nude content of an image. Now, we will show the experimental analysis of some nude images for testing our algorithm. Nude image without face region is rare, in the following section we will study some nude images with faces.

Image 1:

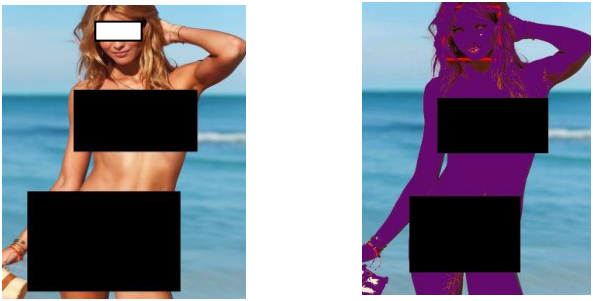

Fig. 2. Image 1, after detect face and skin pixels

Image 2:

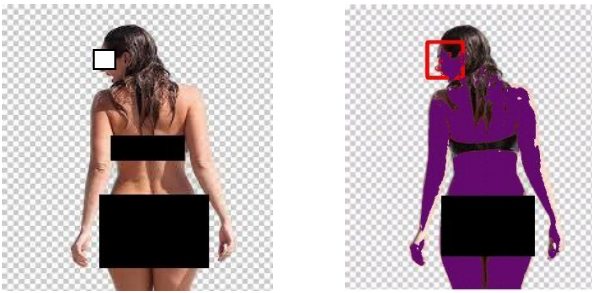

Fig. 3. Image 2, after detect face and skin pixels

Image 3:

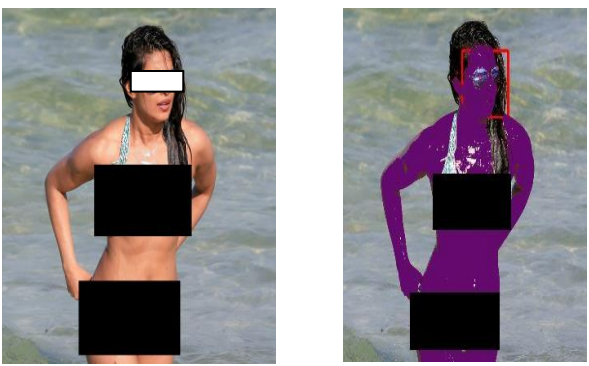

Fig. 4. Image 3, after detect face and skin pixels

Image 4:

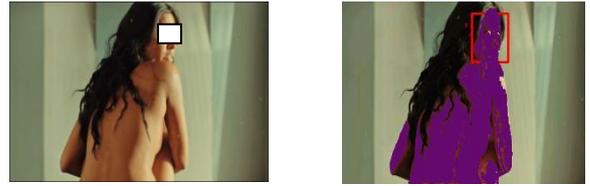

Fig. 5. Image 4, after detect face and skin pixels

Image 5:

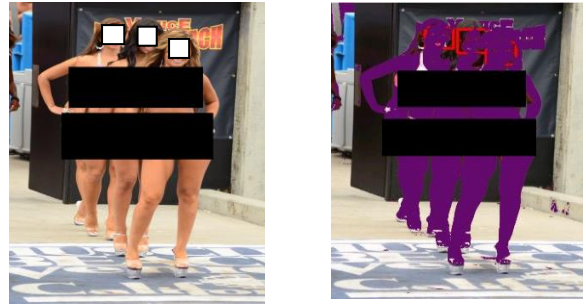

Fig. 6. Image 5, after detect face and skin pixels

TABLE II. NUDITY ANALYSIS OF IMAGES

| Image | Skin pixels | Face region skin pixels | Ratio(b/a) / Nudity | False-positive (skinDetection) |
|---|---|---|---|---|
| Image 1 | 292900 | 23411 | 0.07 / **True** | 5% |
| Image 2 | 67037 | 1875 | 0.02 / **True** | 1% |
| Image 3 | 229369 | 18404 | 0.08 / **True** | 0.03% |
| Image 4 | 109218 | 10790 | 0.09 / **True** | 0.02% |
| Image 5 | 209464 | 12935 | 0.06 / **True** | 3% |

Here, we take five nude images, calculate their skin pixels, and face region. The ratio of face region pixels and total skin pixels of all the testing image is below our nudity threshold of 0.15%. Hence, all the five testing images flagged as nude-image. For skin detection, the false positive rate is also very low. When the false positive in the skin detection method is low, our proposed method can detect image nudity 99% accurately. On average, our nudity algorithm can detect 95% accurately the image nudity.

## V. CONCLUSIONS

This paper presents a face-skin based lightweight and fast-frontend nudity detection technique for web and mobile application. The nudity concept can easily be used in any other system after applying a reliable face detection technique. Two human body parameters skin regions and face-skin regions are used. If the ratio of that parameters is above a threshold, then the input image is flagged as nude-image.

A powerful RGB and HSV color-based skin detection are used to detect skin regions which false-positive rate is low compare to other color-based skin detection method. For detecting face regions, google cloud vision api is used for testing web application images and mobile vision api is used in mobile application images.

From the experimental result, the proposed nudity detection algorithm can detect image nudity 95% accurately with a low false-positive rate.